\documentclass[10pt,twocolumn,letterpaper]{article}

\usepackage{cvpr}              

\usepackage{graphicx}
\usepackage{amsmath}
\usepackage{amssymb}
\usepackage{booktabs}
\usepackage{enumitem}
\setitemize{noitemsep,topsep=0pt,parsep=0pt,partopsep=0pt}

\usepackage[pagebackref,breaklinks,colorlinks]{hyperref}

\usepackage[capitalize]{cleveref}
\crefname{section}{Sec.}{Secs.}
\Crefname{section}{Section}{Sections}
\Crefname{table}{Table}{Tables}
\crefname{table}{Tab.}{Tabs.}

\def\cvprPaperID{7174} 
\def\confName{CVPR}
\def\confYear{2023}

\begin{document}

\title{Inverse Rendering of Translucent Objects using Physical and Neural Renderers}

\author{Chenhao Li, Trung Thanh Ngo, Hajime Nagahara\\
Institute for Datability Science, Osaka University\\
{\tt\small \{lch@is.ids, trung@am.sanken, nagahara@ids\}.osaka-u.ac.jp}
}
\maketitle

\begin{abstract}
   In this work, we propose an inverse rendering model that estimates 3D shape, spatially-varying reflectance, homogeneous subsurface scattering parameters, and an environment illumination jointly from only a pair of captured images of a translucent object. In order to solve the ambiguity problem of inverse rendering, we use a physically-based renderer and a neural renderer for scene reconstruction and material editing. Because two renderers are differentiable, we can compute a reconstruction loss to assist parameter estimation. To enhance the supervision of the proposed neural renderer, we also propose an augmented loss. In addition, we use a flash and no-flash image pair as the input. To supervise the training, we constructed a large-scale synthetic dataset of translucent objects, which consists of 117K scenes. Qualitative and quantitative results on both synthetic and real-world datasets demonstrated the effectiveness of the proposed model. Code and Data are available at \href{https://github.com/ligoudaner377/homo_translucent}{https://github.com/ligoudaner377/homo\_translucent} 
\end{abstract}

\begin{figure*}[t]
  \centering
  \includegraphics[width=\textwidth]{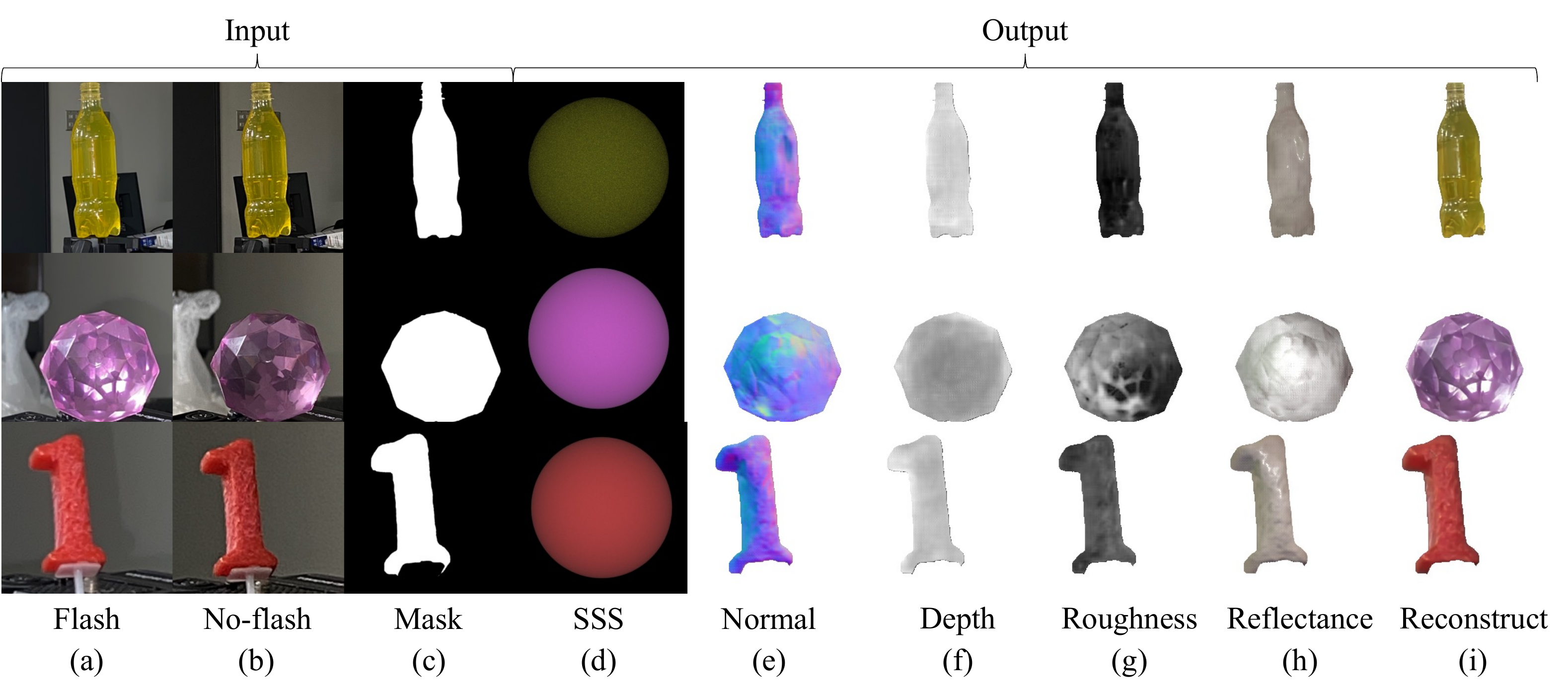}
  \caption{Inverse rendering results of real-world translucent objects. Our model takes (a) a flash image, (b) a no-flash image, and (c) a mask as input. Then, it decomposes them into (d) homogeneous subsurface scattering parameters (denoted as SSS in the figure), (e) surface normal, (f) depth, and (g) spatially-varying roughness. We use the predicted parameters to re-render the image that only considers (h) the surface reflectance, and (i) both surface reflectance and subsurface scattering. Note that (d) subsurface scattering only contains 7 parameters (3 for extinction coefficient $\sigma_t$, 3 for volumetric albedo $\alpha$, 1 for phase function parameter $g$), we use these parameters to render a sphere using Mitsuba \cite{nimier2019mitsuba} for visualization. Brighter areas in the depth map represent a greater distance, and brighter areas in the roughness map represent a rougher surface.}
  \label{fig:real}
\end{figure*}

\section{Introduction}

Inverse rendering is a long-standing problem in computer vision, which decomposes captured images into multiple intrinsic factors such as geometry, illumination, and material. With the estimated factors, many challenging applications like relighting \cite{pandey2021total, liu2020learning, yu2019inverserendernet, sengupta2018sfsnet, xu2018deep, zhou2019deep}, material editing \cite{liu2017material, shi2017learning} and object manipulation \cite{yao20183d, wu2017neural, kulkarni2015deep, zhang2020image} become feasible. In this work, we focus on a very complicated object for inverse rendering: translucent objects. From biological tissue to various minerals, from industrial raw materials to a glass of milk, translucent objects are everywhere in our daily lives. A critical physical phenomenon in translucent objects is SubSurface Scattering (SSS): a photon penetrates the object's surface and, after multiple bounces, is eventually absorbed or exits at a different surface point. This non-linear, multiple bounces, multiple path process makes inverse rendering extremely ill-posed. To reduce the complexity of the task, we assume that the SSS parameters of the object are constant in a continuous 3-dimensional space (homogeneous subsurface scattering). So, given a translucent object, our task is to simultaneously estimate shape, spatially-varying reflectance, homogeneous SSS parameters, and environment illumination. 

However, a well-known issue of inverse rendering is the ambiguity problem—the final appearance of an object results from a combination of illumination, geometry, and material. For instance, a brighter area in the image may be due to the specular highlight. It may be very close to the light source; the coloration may also come from the light source or the object itself. The situation becomes more complicated when SSS is also considered because it is hard to tell the object intensity comes whether from the surface, subsurface, or both.

Solutions to the ambiguity problem can be roughly divided into two groups. The first group of researchers addresses the ambiguity problem by providing more information to the model. For example, some works use multi-view \cite{xu2019deep, yariv2020multiview, bi2020deep, zhang2021physg, li2020through, boss2021nerd, boss2021neuralpil} or multi-light \cite{boss2020two, xu2019deep} set up. The solution of the second group is based on various assumptions and simplifications. For example, some researchers simplify the reflectance model by assuming a Lambertian reflectance \cite{sengupta2018sfsnet}, some simplify the illumination by assuming that the object is illuminated by a single point light source \cite{deschaintre2018single}, some simplify the geometry by assuming a near-planar \cite{li2018materials}. For the SSS, most existing works \cite{barron2014shape, li2018learning, sengupta2019neural, boss2020two, sang2020single, li2020inverse, wu2021rendering, nam2018practical, deschaintre2021deep, Boss2019-SingleImageBrdf, chen2021dib, wimbauer2022rendering} simply ignore it by assuming that light does not penetrate the object's surface, and the final appearance only depends on the pure surface reflectance. On the other hand, some researchers \cite{levis2015airborne, gkioulekas2016evaluation, khungurn2015matching, zhao2016downsampling, gkioulekas2013inverse, che2020towards} consider a pure SSS without surface reflectance. Some works \cite{dong2014scattering, inoshita2014surface, deng2022reconstructing, yang2016inverse} use the BSSRDF model to simplify SSS of optically thick materials in the form of complex surface reflectance. The limitation of these works is obvious: most translucent objects (\textit{e.g.} wax, plastic, crystal) in our world can easily break their assumptions, as it contains both surface reflectance and SSS.
We tackle a more challenging problem by considering both surface reflectance and SSS of translucent objects with an arbitrary shape under an environment illumination. However, using a complex scene representation means introducing more ambiguity. 

In this work, we propose an inverse rendering framework that handles both surface reflectance and SSS. Specifically, we use a deep neural network for parameter estimation. To enable the proposed model to train and estimate the intrinsic physical parameters more efficiently, we employ two differentiable renderers: a physically-based renderer that only considers the surface reflectance of the direct illumination and a neural renderer that creates the multiple bounce illumination as well as SSS effect. Two renderers work together to re-render the input scene based on the estimated parameters and at the same time enable material editing. To enhance the supervision of the proposed neural renderer, we also propose an augmented loss by editing the SSS parameters. Moreover, inspired by the recent BRDF estimation methods \cite{boss2020two, aittala2015two} that use a flash and no-flash set up to address the problem of unpredictability in saturated highlight, we also adopt this two-shot setup not only to deal with the saturated highlight problem but also to disentangle the surface reflectance and SSS. To train our model, we construct a synthetic dataset consisting of more than 117K translucent scenes because there is no sufficient dataset supporting translucent objects. Each scene contains a human-created 3D model and is rendered with a spatially-varying microfacet BSDF, homogeneous SSS, under an environment illumination.

Our contributions are summarized as follows:
\begin{itemize}
\item We first tackle the problem of estimating shape, spatially-varying surface reflectance, homogeneous SSS, and illumination simultaneously from a flash and no-flash pair of captured images at a single viewpoint.
\item We build a novel model that combines a physically-based renderer and a neural renderer for explicitly separating the SSS and the other parameters.
\item We introduce the augmented loss to train the neural renderer supervised by altered images whose SSS parameters were edited. 
\item We construct a large-scale photorealistic synthetic dataset that consists of more than 117K scenes. 
\end{itemize}

\section{Related Work}
\label{sec:related work}

\subsection{Inverse rendering of surface reflectance}
There has been a lot of work on estimating depth \cite{ranftl2019towards,fu2018deep,li2018deep}, BRDF \cite{aittala2016reflectance, aittala2015two, li2018materials, li2017modeling, deschaintre2018single, meka2018lime}, and illumination \cite{gardner2017learning, georgoulis2017around, hold2017deep} separately. 

With the recent rise of deep learning, more and more research is focusing on simultaneous parameter estimation. Li \etal \cite{li2018learning} propose the first single-view method to estimate the shape, SVBRDF, and illumination of a single object. They use a cascaded network to tackle the ambiguity problem and an in-network rendering layer to create global illumination. A few research groups demonstrate that a similar idea can be applied in more complex situations, such as indoor scene inverse rendering by using a Residual Appearance Renderer \cite{sengupta2019neural} or spatially-varying lighting \cite{li2020inverse, Wang_2021_ICCV, zhu2022irisformer}. Li \etal \cite{li2022phyir} extend the inverse rendering of indoor scenes by considering more complex materials like metal or mirror. Boss \etal \cite{boss2020two} tackle the ambiguity problem under a saturated highlight by using a two-shot setup. Sang \etal \cite{sang2020single} implement shape, SVBRDF estimation, and relighting by using a physically-based renderer and a neural renderer. Deschaintre \etal \cite{deschaintre2021deep} introduce polarization into shape and SVBRDF estimation. Wu \etal \cite{wu2021rendering} train the model in an unsupervised way with the help of the rotational symmetry of specific objects. Lichy \etal \cite{lichy2021shape} enables a high resolution shape and material reconstruction by a recursive neural architecture. Our model can be interpreted as an extended version of these methods that also take SSS into account.

\subsection{Inverse rendering of subsurface scattering}

Subsurface scattering is essential for rendering translucent objects such as human skin, minerals, smoke, etc. The reconstruction and editing of these objects heavily rely on scattering parameter estimation. However, inverse scattering is a challenging task due to the multiple bounces and multiple paths of light inside the object. Some works \cite{levis2015airborne, gkioulekas2016evaluation, khungurn2015matching, zhao2016downsampling, gkioulekas2013inverse} combine analysis by synthesis and Monte Carlo volume rendering techniques to tackle this problem. However, they suffer from some problems like local minimal and long optimization time.

Recently, Che \etal \cite{che2020towards} used a deep neural network to predict the homogeneous scattering parameters as a warm start for analysis by synthesis. However, they do not estimate parameters such as geometry, illumination, etc. This means that their model can not handle applications like scene reconstruction, material editing. In addition, they only consider pure SSS. However, most translucent objects in our real-world consist of both surface reflectance and SSS. Our model estimates geometry, illumination, surface reflectance, and SSS simultaneously. 

\subsection{Differentiable rendering}

Differentiable renderers are broadly used in inverse rendering field for reconstructing human faces \cite{sengupta2018sfsnet, wen2021self}, indoor scenes \cite{sengupta2019neural, li2020inverse}, buildings \cite{ meshry2019neural, yu2019inverserendernet}, and single objects \cite{li2018learning, boss2020two, sang2020single}. However, most of them use the physically-based renderer that only considers direct illumination. Such a renderer cannot produce high-quality images as they do not allow for effects like soft shadows, interreflection, and SSS. Recently, there have been some general-purpose differentiable renderers \cite{nimier2019mitsuba, li2018differentiable, Zhang:2019:DTRT, Zhang:2020:PSDR} that consider indirect illumination. However, such a Monte Carlo path tracing graph takes a lot of memory and computing resources, especially when a large sample per pixel (spp) is used. Instead, some works \cite{li2018learning, nalbach2017deep, sengupta2019neural, sang2020single, yu2020self,zhang2021path} splice a neural network behind direct illumination only renderer to enable global illumination or scene editing. Inspired by them, we also designed a renderer-neural network architecture for the SSS task.

\subsection{Scene editing}
Scene editing is a broad research topic, and recent research advances have been dominated by deep learning. A bunch of works has been reported for relighting \cite{pandey2021total, liu2020learning, yu2019inverserendernet, sengupta2018sfsnet, xu2018deep, zhou2019deep}, material editing \cite{liu2017material, shi2017learning} and object manipulation \cite{yao20183d, wu2017neural, kulkarni2015deep, zhang2020image}. We take the first step on the scattering parameter editing of translucent objects by observing only two images.

\begin{figure*}[t]
  \centering
  \includegraphics[width=\linewidth]{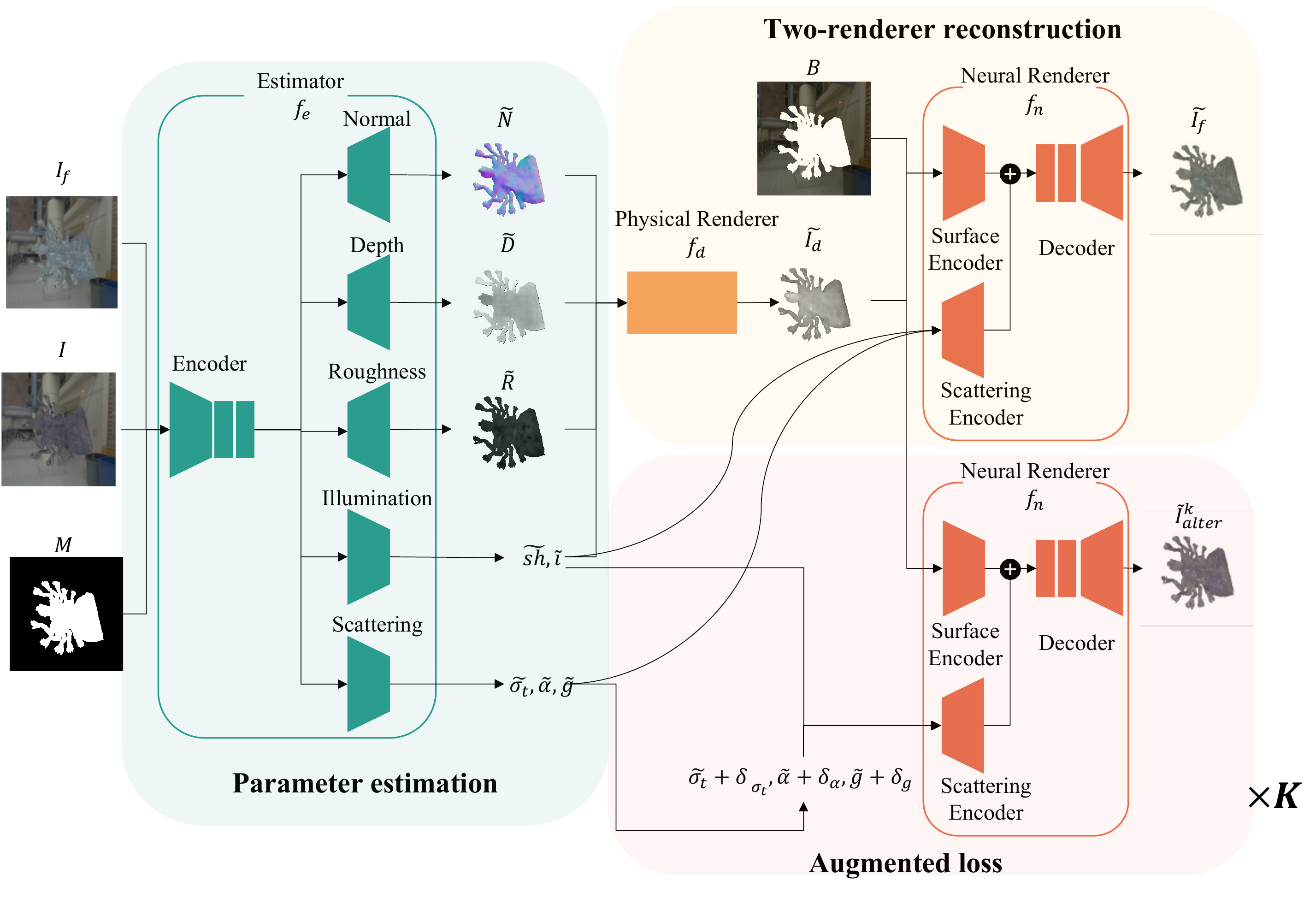}
  \caption{Overview of the proposed model. We use different colors to indicate different functions: green for the estimator, orange for the physically-based renderer, red for the neural renderer. We have $K$ augmented loss modules and only show one in the figure.}
  \label{fig:overview}
\end{figure*}

\section{Methods}
In this section, we introduce the proposed inverse rendering framework. Section \ref{sec:psmd} presents how we model our input and output. In Section \ref{sec:esti}, we talk about a neural network that is used for parameter estimation. In Section \ref{sec:two-renderer} we discuss the proposed two-renderer structure. Then, we illustrate an augmented loss to enhance the supervision of the proposed neural renderer in Section \ref{sec:multi-branch}. Finally, in Section \ref{sec:loss} we show the loss function.

\subsection{Problem setup}
\label{sec:psmd}
\textbf{Scene representation} 
For the geometry part, we use a depth map $D$ to roughly represent the shape of the object and a normal map $N$ to provide local details. For the surface part, we use a microfacet BSDF model proposed by \cite{walter2007microfacet}. A roughness map $R$ is used to represent BSDF parameters. The homogeneous SSS is modeled by three terms: an extinction coefficient $\sigma_t$ which controls the optical density, a volumetric albedo $\alpha$ which determines the probability of whether photons are scattered or absorbed during a volume event, and a Henyey-Greenstein phase function~\cite{HGphase1941} parameter $g$ which defines whether the scattering is forward ($g>0$), backward ($g<0$) or isotropic ($g=0$). We estimate the spherical harmonics $sh$ as a side-prediction to assist the other part of the model. A flashlight intensity $i$ is also predicted, taking into account that the intensity of the flashlight varies from device to device.

\textbf{Model design} 
Inspired by Aittala \etal \cite{aittala2015two} and Boss \etal \cite{boss2020two}, we also use a flash and no-flash image setup. Our motivation for this design is that the visibility of translucent objects will vary at different light intensities. For example, if a bright light is placed on the back of a finger, we can clearly see the color of the blood. Such a property facilitates scattering parameter estimation and enables better disentanglement of surface reflectance and SSS. 

So, given a translucent object of unknown shape, material, and under unknown illumination, our target is to estimate these parameters simultaneously. In addition, we also enable material editing by manipulating the estimated parameters. Figure \ref{fig:overview} shows an overview of our model. The inputs of our model are three images: a flash image $I_f  \in \mathbb{R} ^{3\times256\times256}$, a no-flash image $I \in \mathbb{R} ^{3\times 256\times256}$, and a binary mask $M  \in \mathbb{R} ^{256\times256}$. For each scene, the estimated parameters are:
\begin{itemize}
\item A depth map $D \in \mathbb{R} ^{256\times256}$ and a normal map $N  \in \mathbb{R} ^{3\times256\times256}$ to represent the shape. 
\item A roughness map $R  \in \mathbb{R} ^{256\times256}$ used in the microfacet BSDF model.
\item The spherical harmonics $sh  \in \mathbb{R} ^{3\times9}$ and a flashlight intensity $i \in \mathbb{R} ^{1}$ to represent illumination.
\item The extinction coefficient $\sigma_t  \in \mathbb{R} ^{3}$, volumetric albedo $\alpha  \in \mathbb{R} ^{3}$, and Henyey-Greenstein phase function parameter $g  \in \mathbb{R} ^{1}$. 
\end{itemize}

\subsection{Parameter estimation}
\label{sec:esti}

Taking advantage of recent developments in deep learning, we use a deep convolutional neural network as our estimator. We use the one encoder and multiple heads structure. Our motivation for this design is that estimating shape, material, and illumination can be thought of as multi-task learning. Ideally, each task can assist the other to learn a robust encoder. The encoder extract features from the input images and each head estimate parameters accordingly. So, given a flash image $I_f$, no-flash image $I$, and a binary mask $M$, the estimated physical parameters are:
\begin{equation}
\tilde{D},\tilde{N},\tilde{R},\tilde{sh},\tilde{i},\tilde{\sigma_t},\tilde{\alpha},\tilde{g} = f_e(I, I_f, M),
\end{equation}
where $f_e$ denotes the estimator. $\tilde{D},\tilde{N},\tilde{R},\tilde{sh},\tilde{i},\tilde{\sigma_t},\tilde{\alpha},\tilde{g}$ are the estimated depth, normal, roughness, spherical harmonics coefficient, flashlight intensity, extinction coefficient, volumetric albedo, and Henyey-Greenstein phase function parameter.

\subsection{Physical renderer and neural renderer}
\label{sec:two-renderer}
The use of reconstruction loss to supervise network training is a widespread technique in deep learning. Specifically, reconstruction in inverse rendering means re-rendering the scene with the estimated parameters. In addition, the rendering module must be differentiable to pass the gradient of reconstruction loss to the estimator.

One potential option is to use a general-purpose differentiable renderer such as Mitsuba\cite{nimier2019mitsuba}. However, we have two problems. First, current general-purpose differentiable renderers have memory and training speed problems for consumer-grade GPUs, especially when using a high sample rate. Because the computational cost of path tracing is much larger than that of standard neural works. For example, differentially rendering a single translucent object image with $256\times256$ resolution and 64 samples per pixel (spp) requires more than 5 seconds and 20 GB memory on an RTX3090 GPU. Moreover, the rendering time and memory will increase linearly when spp increase. Second, re-rendering the object with SSS requires the entire 3D shape (\textit{e.g.}, a 3D mesh). However, it is not easy to estimate the complete 3D information at a single viewpoint. That is the reason we only estimate a normal and a depth map as our geometry representation. This representation is sufficient for rendering surface reflectance but not for SSS. Another option \cite{wu2021rendering, boss2020two} is to use a differentiable renderer that only considers direct illumination. This sacrifices some reconstruction quality but dramatically improves efficiency. However, such a method cannot be applied to a translucent object because SSS depends on multiple bounces of light inside the object. 

With the assist of recent advances in image-to-image translation \cite{pix2pix2017, zhu2017unpaired}, many works have demonstrated the successful use of neural networks for adding indirect illumination \cite{li2018learning, nalbach2017deep}, photorealistic effect \cite{meshry2019neural, sengupta2019neural}, or relighting\cite{sang2020single, yu2020self}. Inspired by them, we propose a two-step rendering pipeline to mimic the rendering performance of SSS of a general-purpose differentiable renderer. The first step is a physically-based rendering module that only considers direct illumination:
\begin{equation}
\tilde{I_d} = f_d(\tilde{D},\tilde{N},\tilde{R},\tilde{i},\tilde{sh}),
\end{equation}
where $f_d$ is a physically-based renderer that follow the implementation of \cite{sang2020single}, and it has no trainable parameters. $\tilde{I_d}$ is the re-rendered image that only consider the surface reflectance. The second step is a neural renderer $f_n$ to create SSS effect:
\begin{equation}
\tilde{I_f} = f_n(\tilde{I_d},\tilde{sh},\tilde{i},\tilde{\sigma_t},\tilde{\alpha},\tilde{g}, B),
\end{equation}
where $\tilde{I_f}$ is the re-rendered flash image. $f_n$ is the proposed neural renderer, and it consists of 3 parts (See Figure \ref{fig:overview} for reference), a Surface encoder, a Scattering encoder, and a decoder. The Surface encoder maps the estimated surface reflectance image $\tilde{I_d}$ into a feature map. In addition to the surface reflectance image, we also input the background image $B$ (masked-out version of $I_f$) to the Surface encoder to provide high frequency illumination information. The Scattering encoder consists of a few upsampling layers. It maps $\tilde{sh},\tilde{i},\tilde{\sigma_t},\tilde{\alpha},\tilde{g}$ to a feature map. The decoder consists of some Resnet blocks \cite{he2016deep} and upsampling layers. The task of our neural renderer can be considered as a conditional image-to-image translation, where the condition is the SSS parameters.

 The advantage of the two-renderer design is that it is naturally differentiable. At the same time, the training cost is acceptable. In addition, the physically-based renderer can provide physical hints to the neural renderer. Because we separate the reconstruction of surface reflectance and SSS explicitly, the problem of ambiguity can be alleviated.
 
  \subsection{Augmented loss}
 \label{sec:multi-branch}
 In this subsection, we present an augmented loss to address the hidden information problem by using multiple altered images to supervise the proposed neural renderer. In Section \ref{sec:two-renderer}, we propose a two-renderer structure to compute the reconstruction images to improve the parameter estimation. However, a well-known problem of reconstruction loss in deep learning is that neural networks learn to “hide” information within them \cite{chu2017cyclegan}. This may cause our neural renderer to ignore the estimated SSS parameters and only reconstruct the input image by the hidden information. If so, the reconstruction loss cannot give correct gradients and thus fail to guide the training of the estimator.
 
 In order to solve this problem, we enhance the supervision of the proposed neural renderer by an augmented loss. Specifically, after the parameters are estimated, we edit the estimated SSS parameters and let the Neural renderer reconstruct image based on the edited SSS parameters. We also render $K$ altered images $I_{alter}^k$ as their ground truth labels to train the neural renderer. The altered images have the same parameters as the original flash image, except the SSS parameters. Specifically, $I_f$ and $I_{alter}^k$ share the same shape, surface reflectance, and illumination, but different extinction coefficient, phase function, and volumetric albedo. The edited SSS parameters are randomly sampled from the same distribution as the original ones:
\begin{equation}
\tilde{I}_{alter}^k = f_n(\tilde{I_d},\tilde{sh},\tilde{i},\tilde{\sigma_t}+\delta_{\sigma_t},\tilde{\alpha}+\delta_\alpha,\tilde{g}+\delta_g, B),
\end{equation}
where $\delta_{\sigma_t}, \delta_\alpha, \delta_g$ are the differences between target SSS parameters (subsurface scattering parameters of $I_{alter}^k$) and the estimated ones, $\tilde{I}_{alter}^k$ is the reconstructed altered images. The benefits of this design are as follows. First, the input image is not the same as the image to be reconstructed, which makes it meaningless for the neural network to ``hide" the information of the original input image. Second, the variety of input parameters and output images makes the neural renderer more sensitive to the changes in SSS parameters. These two points make the neural renderer a good guide for the estimator, which allows for a more accurate estimation of the SSS parameters. Considering the time and computing resources it takes to render the training images, in practice, we set $K=3$.

\begin{figure*}[t]
  \centering
  \includegraphics[width=\textwidth]{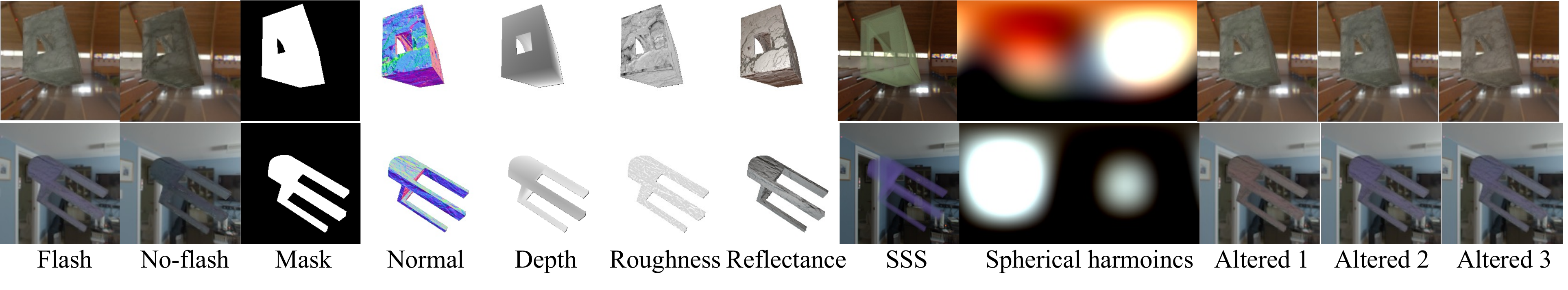}
  \caption{Examples of the proposed dataset, we use the ground truth shape and illumination to visualize SSS parameters.}
  \label{fig:dataset}
\end{figure*}

\subsection{Loss functions}
\label{sec:loss}
The proposed model is fully supervised and trained end to end, and we compute the loss between the estimated parameters their ground truths:
\begin{eqnarray}
L &=& L_D + L_N + L_R + L_{sh} + L_{i} \\ \nonumber
&+& L_{\sigma_t} + L_{\alpha} + L_g + L_{I_f} + \sum_K{L_{alter}^k},
\end{eqnarray}
where $L_D,L_N,L_R,L_{I_f}, L_{alter}^k$ stand for the $L1$ loss between the estimated depth, normal, roughness, flash image, altered images and their ground truth ones.  $L_{sh},L_{i},L_{\sigma_t},L_{\alpha},L_g$ stands for the $L2$ loss between the estimated spherical harmonics, flashlight intensity, extinction coefficient, volumetric albedo, Henyey-Greenstein phase function parameter and their ground truth ones. 

\begin{table*}[t]
  \centering
  \caption{MAE results on 17140 test scenes. For each element we report mean(std) value. The scale of mean is $1\times10^0$, and std is $1\times10^{-3}$.}
  \resizebox{\textwidth}{!}{
  \begin{tabular}{*{9}c}
    \toprule
      &\multicolumn{2}{c}{Geometry}&\multicolumn{1}{c}{BSDF}&\multicolumn{2}{c}{Illumination}&\multicolumn{3}{c}{SSS}\\
      &$N$ &$D$ &$R$  &$sh$ &$i$ &$\sigma_t$ &$\alpha$ &$g$ \\
    \midrule
    Baseline &.0918(.4395)	&.0705(.4443)	&.0811(.5303)	&.1083(.6571)	&.0912(1.042)	&.1670(.7904)	&.1061(.5792)	&.1762(.8811)\\
    2R       &.0916(.3009)	&.0697(.3617)	&.0811(.4903)	&.1064(.8984)	&.0908(.7697)	&.1675(1.072)	&.1057(.4191)	&.1777(2.562)\\
    2R-AUG    &.0913(.1768)	&.0699(1.271)	&.0807(.2714)	&.1105(8.926)	&.0893(1.151)	&.1619(.9635)	&.1040(.1578)  & .1703(.8856)\\
    Che \etal \cite{che2020towards} & - & - & - & - & - & .1828  & .1115 & .2123\\ 
    \textbf{Full model} &\textbf{.0894}(.1532)	&\textbf{.0646}(.3283)	&\textbf{.0769}(.2659)	&\textbf{.0989}(1.017)	&\textbf{.0804}(1.736)		&\textbf{.1590}(.2286) &\textbf{.1002}(.5185)&\textbf{.1655}(.3932)\\
    \bottomrule
  \end{tabular}
  }
  \label{tab:ablation}
\end{table*}

\begin{figure*}[t]
  \centering
  \includegraphics[width=\linewidth]{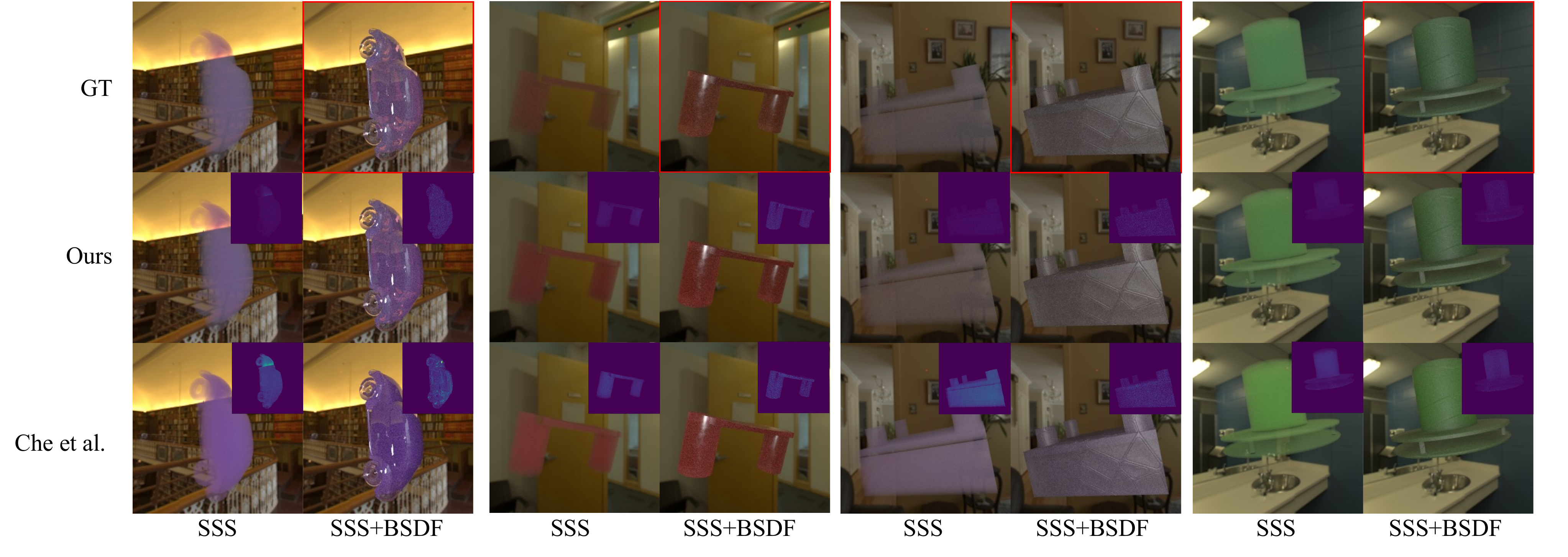}
  \caption{Visual comparison with Che \etal \cite{che2020towards}. We use two visualizations which are ``SSS" and ``SSS+BSDF" to show the parameter estimation results. ``SSS" is rendered by the GT illumination, shape, and predicted SSS. ``SSS+BSDF" is rendered by the GT illumination, shape, BSDF, and predicted SSS. Images in red rectangles are input images. Error maps are in the upper right corner.}
  \label{fig:compare_che}
\end{figure*}

\begin{figure*}[t]
  \centering
  \includegraphics[width=\linewidth]{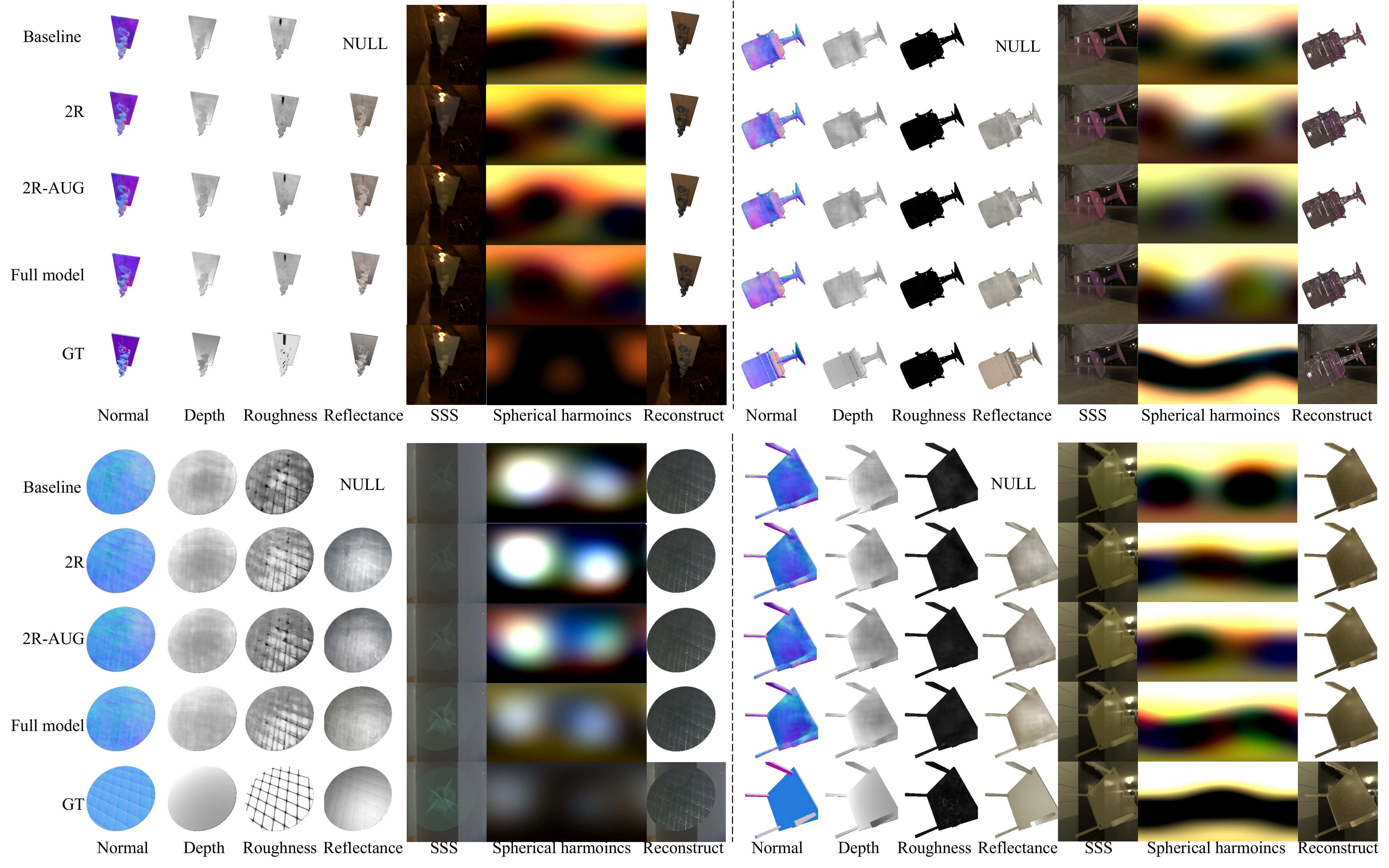}
  \caption{Visual results of ablation study on the synthetic dataset. For visualization, we use the ground truth shape and illumination to render the image of the estimated SubSurface Scattering parameters (denoted as SSS in the figure).}
  \label{fig:synthetic_1}
\end{figure*}

\begin{figure*}[t]
  \centering
  \includegraphics[width=\linewidth]{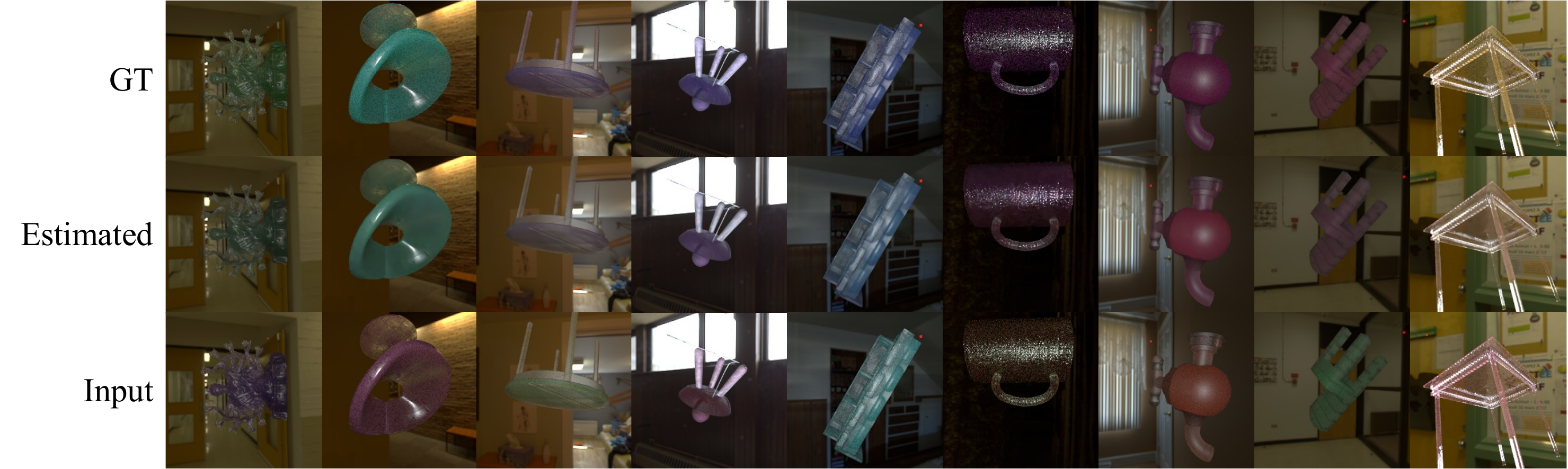}
  \caption{SSS parameter editing results. The 3rd row is the input images. We randomly edit the scattering parameters before inputting them into the neural renderer. The 2nd row is the estimated images, and we show their ground truth images in the 1st row.}
  \label{fig:edit_albedo}
\end{figure*}

\section{Experiments}
We introduce our dataset in Section \ref{sec:datasets}. In Section \ref{sec:comparison}, we compare our model with Che \etal \cite{che2020towards}. We conduct an ablation study and report the quantitative and qualitative results in Section \ref{sec:ablation}.  In Section \ref{sec:material editing} we show the results of SSS parameter editing application. More experiment results can be found in the supplementary document.

\subsection{Datasets}
\label{sec:datasets}
In this work, we propose the first large-scale synthetic translucent dataset that contains both surface reflectance and SSS. We collected the 3D objects from ShapeNet \cite{shapenet2015}, human-created roughness map and auxiliary normal map from several public resources, and environment map from the Laval Indoor HDR dataset \cite{gardner2017learning}. 

For each scene, we used Mitsuba \cite{nimier2019mitsuba} to render five images: flash image, no-flash image, and three altered images. Each scene contains a randomly selected object, auxiliary normal map, roughness map, environment map. We also randomly sample the SSS parameters and flash light intensity. Similar to some SfP (Shape from Polarization) works \cite{ba2020deep}, we assume a constant IoR (Index of Refraction) to reduce the ambiguity problem. A full introduction to our dataset can be found in the supplementary documentation.

\subsection{State-of-the-art comparison}
\label{sec:comparison}
     To the best of our knowledge, we are the first to tackle the inverse rendering problem of translucent objects containing both surface reflectance and SSS. It is not easy to compare our method with \textbf{pure surface reflectance} methods \cite{li2018learning, boss2020two}. Although similar to our method, these methods also predict parameters like surface normal, depth, and roughness. The key problem is that the coloration of pure surface reflectance models is affected by ``diffuse albedo", which is a parameter of BRDF. However, in our proposed scene representation, the surface reflection and refraction are modeled by BSDF, and the SSS is modeled by the Radiative Transport Equation (RTE), which means that the coloration is affected by the volumetric albedo, extinction coefficient, and phase function. In conclusion, training pure surface reflectance methods on our dataset is impossible. Thus, we choose a \textbf{pure SSS} method proposed by Che \etal \cite{che2020towards} to compare with our model. Their method requires an edge map as the additional input of the neural network. We follow their paper to generate the edge maps for our dataset. We train their model on our dataset with the same hyperparameters as our method. We report the Mean Absolute Error (MAE) in Table \ref{tab:ablation}, and visual comparison in Figure \ref{fig:compare_che}. It is difficult to compare the parameter estimation accuracy without GT parameter references, so we only show the synthetic data results. It is observed that their method fails to estimate reasonable SSS parameters due to the highly ambiguous scene representation.

\subsection{Ablation study}
\label{sec:ablation}
We conduct ablation studies to evaluate the effects of each component of the proposed model. Specifically, we start with a Baseline model, which uses the same estimator as our Full model, but only reconstructs the input scene with a neural network. So, the baseline model is essentially an autoencoder. Then, we divide the reconstruction into two steps: a physically-based renderer that reconstructs the surface reflectance and a neural renderer that create the multi-bounce illumination as well as the SSS effect. We call this model ``2R". After that, we introduce the augmented loss to the ``2R" model and denote it as ``2R-AUG". Finally, we implement the Full model by adding the two-shot setting to ``2R-AUG". We report the MAE results on the synthetic data of all experiments in Table \ref{tab:ablation}. It is observed that for most of the metrics, ``2R" model outperforms the Baseline model. Especially for the illumination part, the accuracy improved a lot. This confirms our previous point that is explicitly separating the surface reflectance, and SSS reduces ambiguity. Although the performance of ``2R" and ``2R-AUG" are similar for geometry, illumination, and surface reflectance parts, the result of SSS parameters which are enhanced by the augmented loss, is improved. Comparison between the ``2R-AUG" and Full model proves that the proposed two two-shot problem settings can further reduce the ambiguity problem. Figure \ref{fig:synthetic_1} demonstrate some visual comparisons. It is observed that the SSS images of the Full model are most similar to the ground truth ones, and the other models are unstable. In addition, because of the two-shot setting, the full model's estimation of environment illumination is also more accurate. To show the flexibility of the proposed model, we also test on several common real-world translucent objects, and illustrate the results in Figure \ref{fig:real}. The results show that our model can estimate reasonable SSS parameters. 

\subsection{Material editing} 
\label{sec:material editing}
 Given a translucent object with an unknown shape, illumination, and material, we show that the proposed inverse rendering framework can edit translucent objects based on the given SSS parameters. Figure \ref{fig:edit_albedo}  shows the editing results.

 \section{Conclusions and Limitations}
In this paper, we made the first attempt to jointly estimate shape, spatially-varying surface reflectance, homogeneous SSS, and illumination from a flash and no-flash image pair. The proposed two-shot, two-renderer, and augmented loss reduced the ambiguity of inverse rendering and improved the parameter estimation. In addition, we also constructed a large-scale synthetic dataset of fully labeled translucent objects. The experiments of the real-world dataset demonstrated that our model could be applied to images captured by a smartphone camera. Finally, we demonstrated that our pipeline is also capable of SSS parameter editing.

The proposed method still has some limitations. First, to reduce the ambiguity problem, we set the IoR to be constant and do not estimate it. However, IoR affects light transmission through object boundaries and thus influences the SSS. For some materials with large or small IoR, our model may not work. Second, our model does not support relighting and novel-view synthesis. Unlike pure surface reconstruction models that the estimated normals can be easily applied to physically-based renderers for relighting or novel-view synthesis, rendering translucent objects requires a full 3D estimation (\eg, mesh) including the backside information. However, estimating the complete geometry is difficult for single-view reconstruction method. Solving these challenges can be good future work.
{\small
\bibliographystyle{ieee_fullname}
\bibliography{egbib}
}

\end{document}


\title{Supplementary Document for \\Inverse Rendering of Translucent Objects using Physical and Neural Renderers}  

\maketitle
\thispagestyle{empty}
\appendix


\section{Introduction}
In this document, we provide the supplementary material of the proposed model. In Section \ref{sec:add_res}, we demonstrate the additional experiment results. In Section \ref{sec:datasets}, we introduce the proposed dataset. In Section \ref{sec:scene_representation}, we introduce our scene representation. Section \ref{sec:model_detail} shows the training details of the proposed model. Section \ref{sec:network_structure} demonstrates our network structure.

\begin{figure*}[t]
  \centering
  \includegraphics[width=0.91\linewidth]{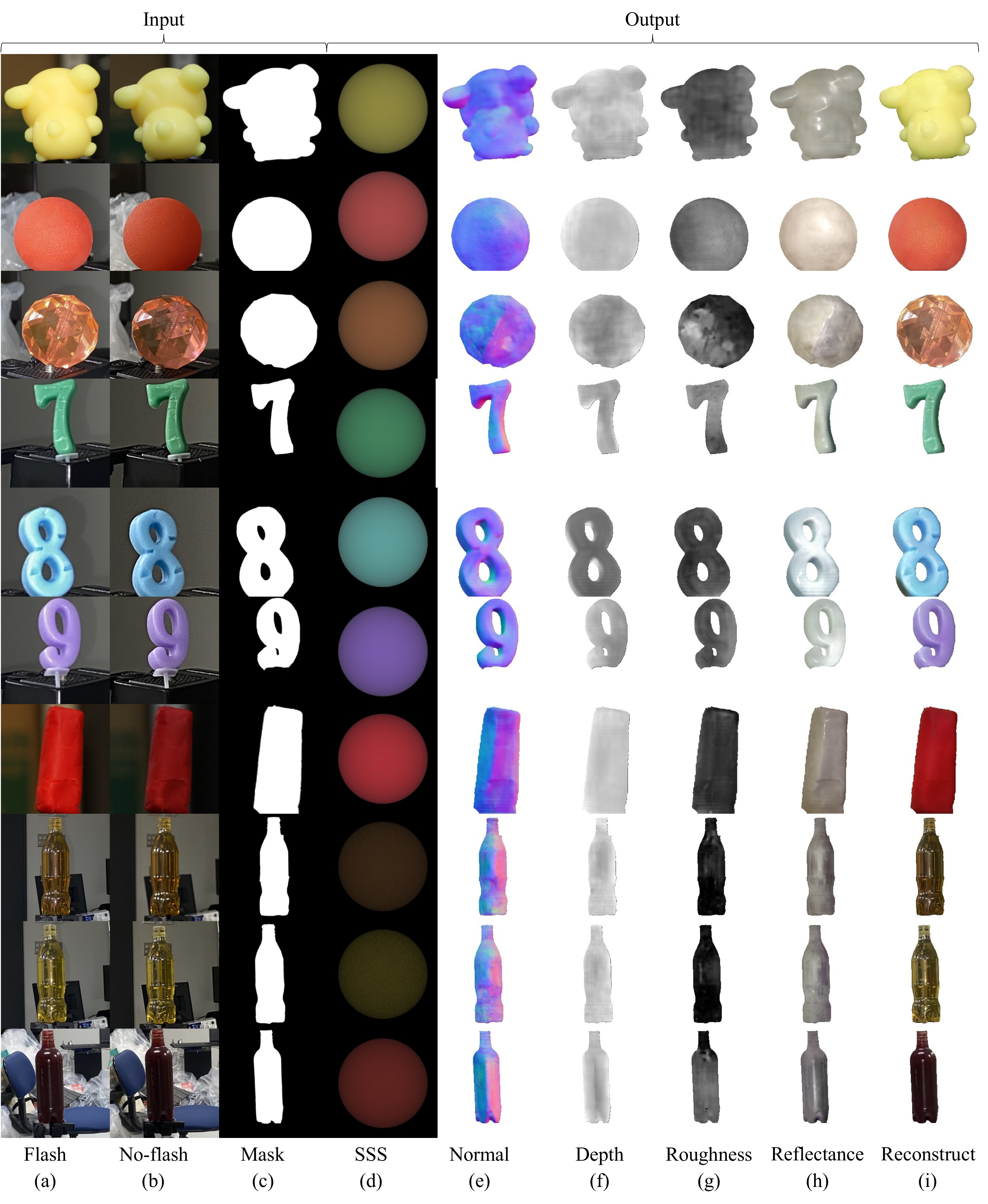}
  \caption{Inverse rendering results of real-world translucent objects. Our model takes a (a) flash image, a (b) no-flash image, and a (c) mask as input. Then, it decomposes them into (d) homogeneous subsurface scattering parameters (denoted as SSS in the figure), (e) surface normal, (f) depth, and (g) spatially-varying roughness. We use the predicted parameters to re-render the image that only considers (h) the surface reflectance, and (i) both surface reflectance and subsurface scattering. Note that (d) subsurface scattering only contains 7 parameters (3 for extinction coefficient $\sigma_t$, 3 for volumetric albedo $\alpha$, 1 for phase function parameter $g$), we use these parameters to render a sphere using Mitsuba2 \cite{nimier2019mitsuba} for visualization. Brighter areas in the depth map represent a greater distance, and brighter areas in the roughness map represent a rougher surface.}
  \label{fig:real_2}
\end{figure*}

\begin{figure*}[t]
  \centering
  \includegraphics[width=0.8\linewidth]{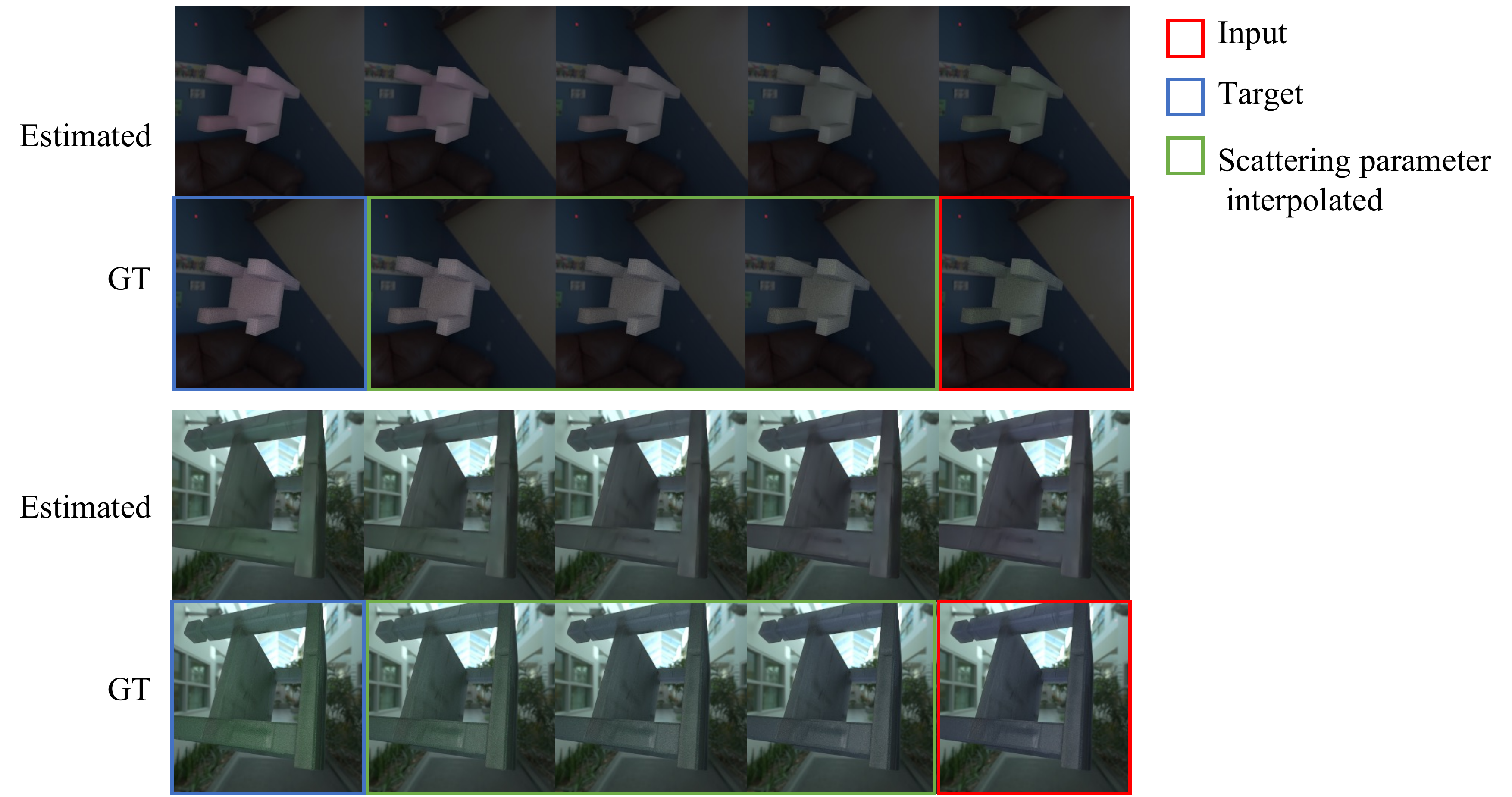}
  \caption{Linear interpolation results of volumetric albedo $\alpha$.}
  \label{fig:interpolation}
\end{figure*}

\begin{figure*}[t]
  \centering
  \includegraphics[width=\linewidth]{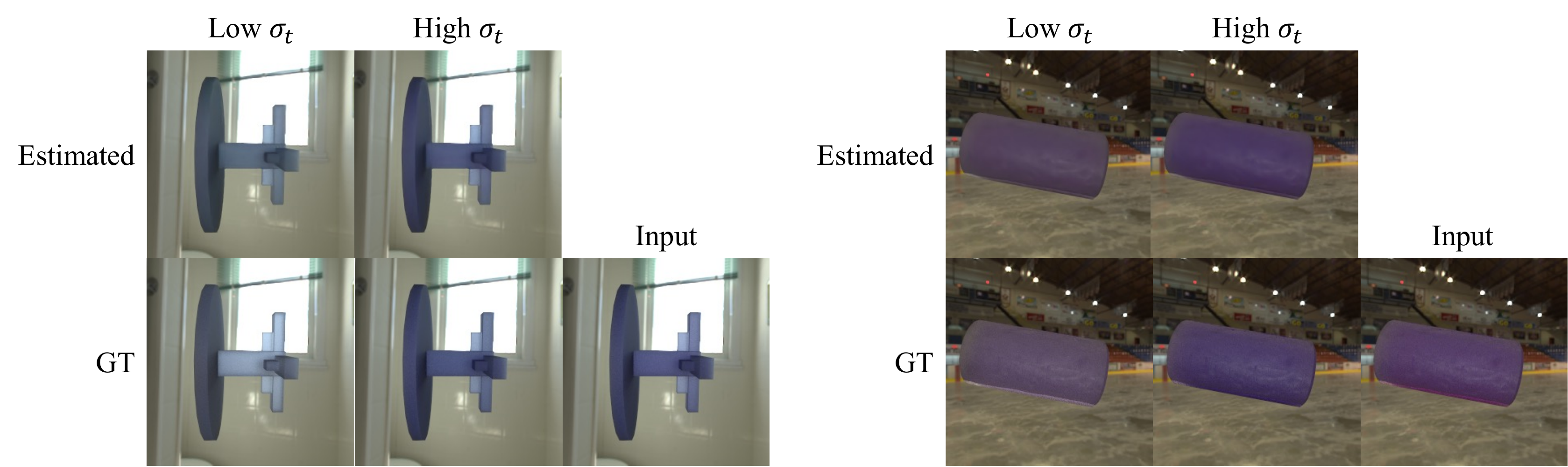}
  \caption{Extinction coefficient $\sigma_t$ editing results. We edit the $\sigma_t$ of the input translucent object and compare the estimated images with their ground truth at two different levels.}
  \label{fig:edit_sigma_t}
\end{figure*}

\begin{figure*}[t]
  \centering
  \includegraphics[width=0.8\linewidth]{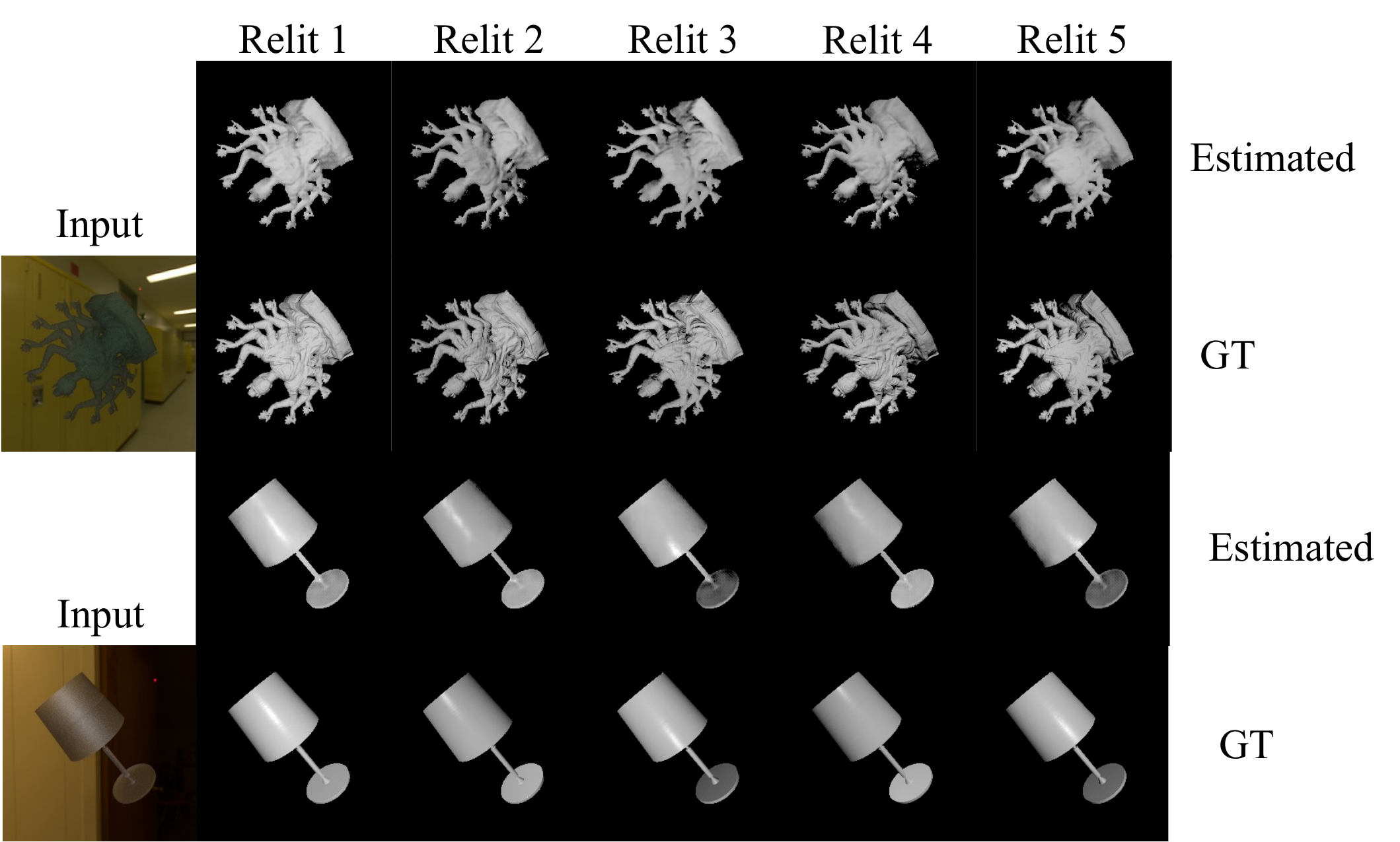}
  \caption{Surface reflectance under different illumination conditions.}
  \label{fig:relit}
\end{figure*}

\begin{figure*}[t]
  \centering
  \includegraphics[width=0.9\textwidth]{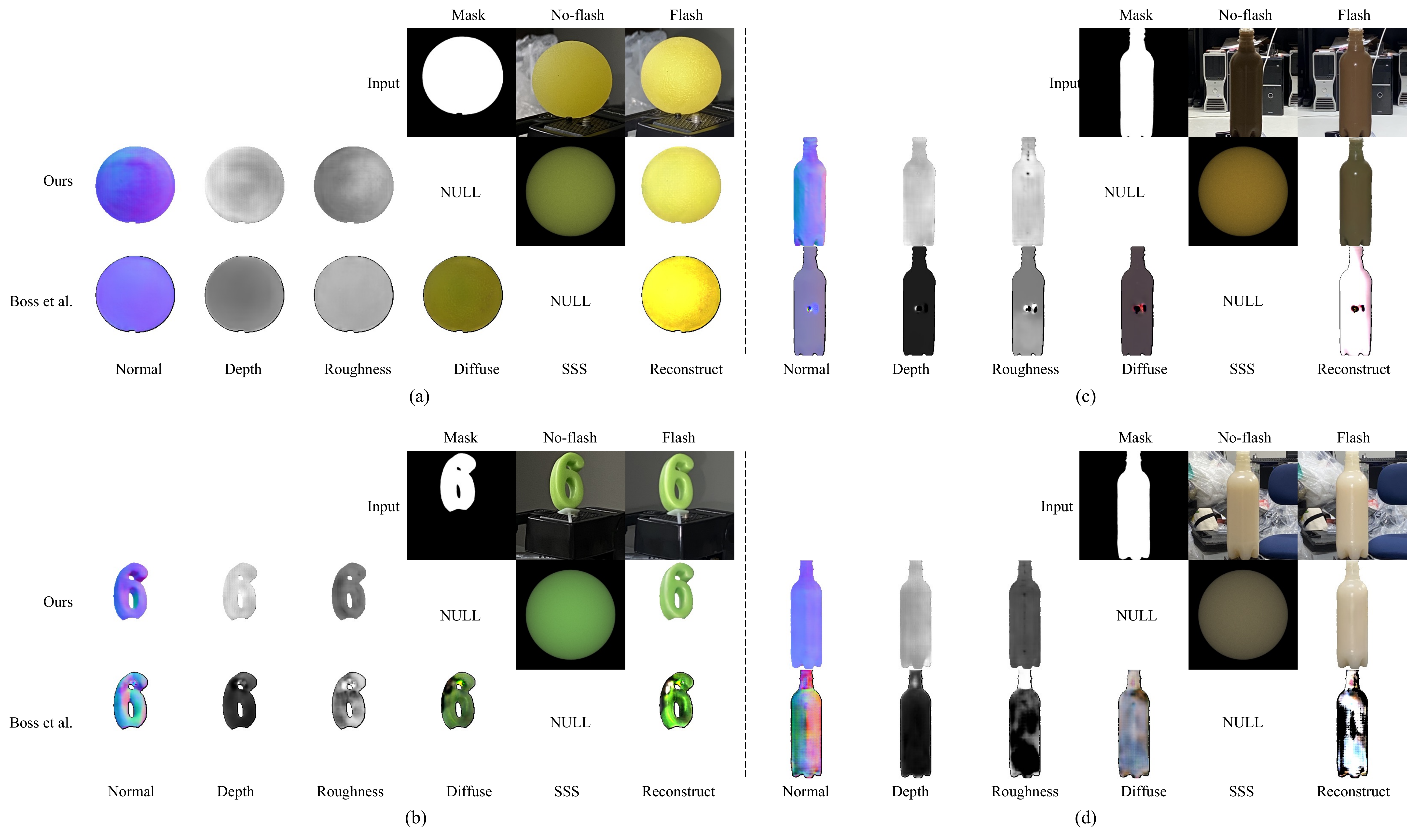}
  \caption{Visual comparison with Boss \etal \cite{boss2020two}. ``Diffuse" in the figure means the surface diffuse albedo. ``SSS" denote the subsurface scattering parameters, we use these parameters to render a sphere using Mitsuba2 \cite{nimier2019mitsuba} for visualization. We select two translucent object (a) and (b) with low transparency, and two translucent object (c) and (d) with high transparency for qualitative comparison. Boss \textit{et al.} \cite{boss2020two} method only works well for low transparency when subsurface scattering is close to surface scattering (diffuse reflectance). Meanwhile, the proposed method works well for both cases.}
  \label{fig:compare}
\end{figure*}

\section{Additional Results}
\label{sec:add_res}
\subsection{Real-world objects decomposition} We demonstrate the additional results of real-world translucent objects in Figure \ref{fig:real_2}. From the figure, we can obseve that our model estimates reasonable subsurface scattering parameters (d) and the re-rendered images (i) look similar to the original input ones (a).

\subsection{Parameter space exploring}
In this section, we explore how well the neural renderer learned the subsurface scattering parameter space by two experiments. The first one is the linear interpolation of volumetric albedo $\alpha$. Given a target volumetric albedo and an original one, we linearly interpolate them and input to our neural renderer. Figure \ref{fig:interpolation} shows that the generated images are similar to their GTs. Second, we use the same method to edit the extinction coefficient $\sigma_t$ from a extremely large value to a small value. We demonstrate the results in Figure \ref{fig:edit_sigma_t}.

\subsection{Relighting of surface reflectance}
Although relighting is not one of our target applications, it is easy to understand how well the normal and depth are estimated by showing the results under novel illumination coditions. We demonstrate them in Figure \ref{fig:relit}. It can be observed that although our model fails to capture some high frequency details of surface normals, it achieves overall good performance.

\subsection{Comparison with the existing works}
 We compare our model with a surface reflectance only method that also uses a two-shot setup proposed by Boss \etal \cite{boss2020two}. They use a novel stage-wise neural network for the shape and SVBRDF estimation. However, if we compare our model with the pure surface reflectance model, a problem arises. For a fair and equitable comparison, all models' training and testing data must be the same. However, training their model requires the GT value of diffuse albedo, which is not a part of the translucent object parameters. So we choose to use their own dataset to train the model and not compare diffuse albedo during testing. We illustrate the results of visual comparison of real-world translucent objects in Figure \ref{fig:compare}. Considering we do not have the same problem setting as their work, the result here is just for user reference.

\section{Datasets}
\label{sec:datasets}
Measuring a large number of real-world translucent objects that include shape, surface reflectance, and subsurface scattering parameters is time-consuming. However, training data is an indispensable part of deep neural networks. Thus, we created a large-scale synthetic dataset by photorealistic rendering. In the following few subsections, we discuss how we prepared the assets for rendering, including 3D objects, BSDF maps, subsurface scattering parameters, and illumination.

\textbf{3D Objects} Some existing works \cite{li2018learning, boss2020two, sang2020single} use the Domain Randomized method to synthesize 3D objects by randomly assembling some simple shapes such as spheres, cylinders, cones, etc. For the shape complexity and diversity of our dataset, we collected human-created 3D objects from ShapeNet \cite{shapenet2015} and some other public resources. However, some objects in the ShapeNet have flipped surface normal, which can result in entirely black pixels when intersecting with light. We used a script to delete these objects, and finally, 5,847 3D objects were retained. All objects were scaled into a cube with a length of 50cm and placed at the origin and were randomly rotated, scaled, and translated during the rendering process. We use 5,000 objects for training and 857 for testing.

\textbf{Roughness and auxiliary normal maps} To make a meaningful surface reflectance pattern, we collected a large number of human-created surface reflectance maps from several open-source websites. Each of them contains a normal auxiliary map and a roughness map. We use a ``smart uv mapping" function of Blender \cite{Hess:2010:BFE:1893021} to apply the collected roughness map and normal maps to objects. The normal map was applied on the original 3D object to modify its surface normal that makes more realistic surface. Before being applied to the objects, all roughness maps and normal maps were randomly resized. In the end, we achieved 2,745 surface reflectance maps and used 2,470 for training and 275 for testing.

\textbf{IoR} In order to reduce the ambiguity problem of inverse rendering, we set the index of refraction (IoR) to a constant (1.5046) and do not estimate it. We choose this value because many of our world's popular materials like glass (1.5046), amber (1.55), and polyethylene (1.49) have similar IORs.

\textbf{Subsurface scattering parameters}
Follow the previous work \cite{che2020towards}, we randomly sampled the subsurface scattering parameters from a uniform distribution with extinction coefficient $\sigma_t \in [0, 32]$, volumetric albedo $\alpha \in [0.3, 0.95]$, and Henyey-Greenstein phase function parameter $g \in [0, 0.9]$.

\textbf{Illumination} To mimic the completed light condition in our real world, we choose environment maps as the light source. We used the Laval Indoor HDR dataset \cite{gardner2017learning}, which consists of 2,357 high-resolution indoor panoramas. During rendering, the pitch and roll were fixed, and they only rotated around the yaw axis. We computed a $3\times9$ Spherical Harmonics coefficients for each environment map to supervise our model. In total, 1500 environment maps were used for training, and 857 were used for testing. 

Initially, we were going to use a point light to simulate the flashlight. However, for some very smooth objects, we observed severe noise when using a point light. Therefore, we used a tiny sphere area light with a radius of 10cm instead of a point light. The area light is placed 10cm behind the camera. Taking into account the different flashlight intensity of various devices in the real world, we also used a random radiance between 35 to 75 $\mathrm{W\cdot m^{-2} sr^{-1}}$.

\textbf{Synthetic dataset} For each scene, we used Mitsuba2 \cite{nimier2019mitsuba} to render five images: flash image, no-flash image, and three altered images. Each scene contains a randomly selected object, normal map, roughness map, environment map, and subsurface scattering parameters. The camera is placed 70cm away from the origin on the positive z-axis and looks at the origin. For the flash image, we used a small area light source behind the viewpoint to simulate the camera flashlight. For the no-flash image, we remove the small area light source and slightly change the camera's look-at direction to simulate the camera shake. For the three altered images, the subsurface scattering parameters were edited from those of the original flash image. In addition, we also rendered the ground truth depth, normal, roughness, and binary masks for the intermediate supervision. We obtained a total of 100,000 training scenes and 17,140 test scenes. 

\textbf{Real-world dataset}
For a more comprehensive test of our model, we also constructed a real-world dataset consisting of several common translucent objects. We took a flash photo and a no-flash photo with a smartphone camera. We manually created a binary mask for each object.

\begin{figure*}[t]
  \centering
  \includegraphics[width=0.8\textwidth]{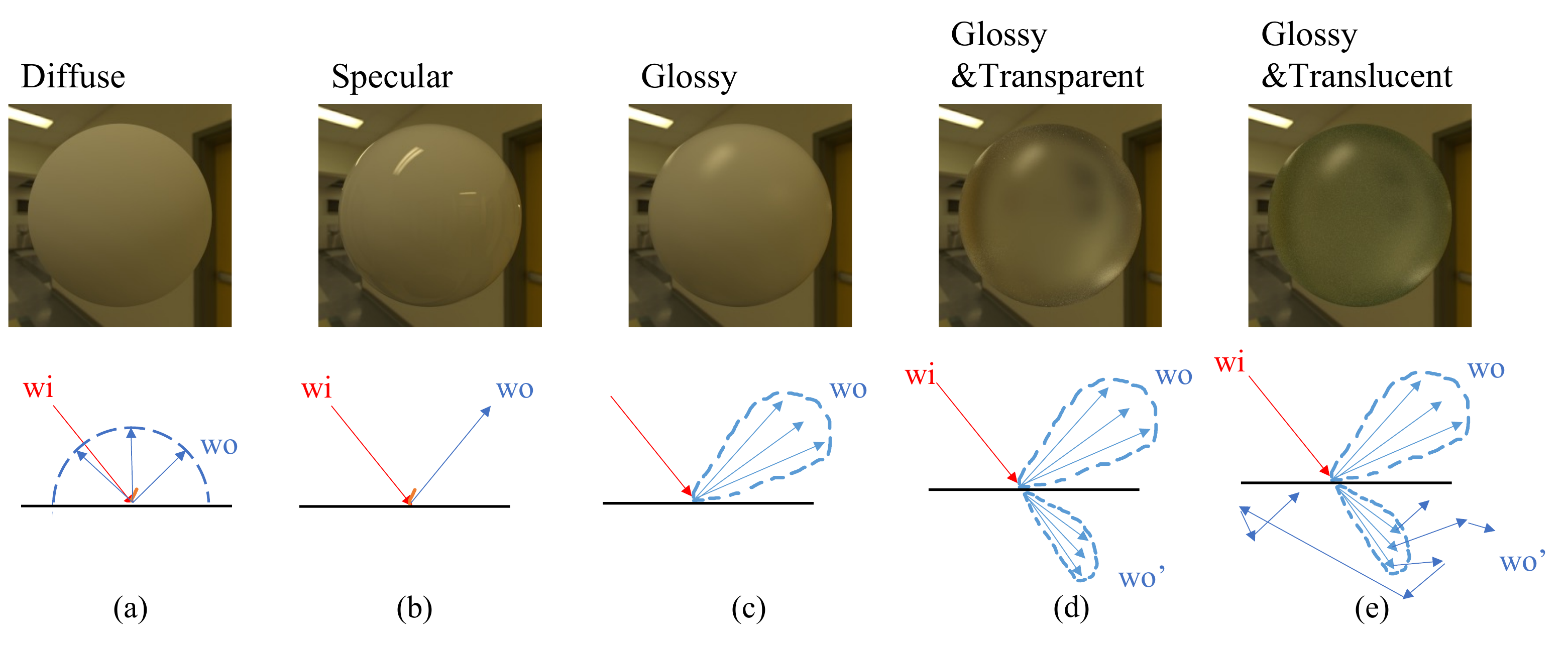}
  \caption{Illustration of different scene representations. The diffuse material (a) assumes that light is reflected uniformly, while the specular material (b) assumes that the incident light and the outgoing light are symmetrical along the surface normal. The glossy material (c) is a generalized version of (a) and (b) that assumes the outgoing light follows a distribution. Introducing refraction into (c) results along a glossy transparent material (d). However, (d) assumes that the light travels in a straight line inside an object. We focus the glossy translucent material (e) which introduces the multiple bounces and multiple paths inside the object.}
  \label{fig:scene_representation}
\end{figure*}

\section{Scene Representation}
\label{sec:scene_representation}
In this section, we introduce the scene representation of our model. Figure \ref{fig:scene_representation} illustrates the differences between our scene representation and the existing works. The simplification of the scene representation can greatly reduce the problem of ambiguity, thereby reducing the difficulty of parameter estimation. However, at the same time, the scenarios in which the model can be applied are also reduced. For example, if we assume a diffuse reflectance like (a) in Figure \ref{fig:scene_representation}, we can only solve the inverse rendering problem of objects like paper, rubber, etc. We make the first attempt at an inverse rendering problem involving both surface reflectance and subsurface scattering. Most translucent objects in our world like wax, plastic, and jade satisfy this scene representation.

For the surface part, we use a microfacet BSDF model proposed by \cite{walter2007microfacet}. Let $p$ be a point at the object surface, the outgoing light radiance $L(w_o)$ of direction $w_o$ at point $p$ is defined by:
\begin{equation}
\begin{split}
&L(w_o) = \\
&\int_\Omega {L(w_i)f_r(w_i, w_o, R(p), N(p))\max(w_i \cdot N(p), 0) \, dw_i} +\\
&\int_{\Omega^{'}} {L(w_i^{'})f_t(w_i^{'}, w_o, R(p), N(p)) \max(w_i^{'} \cdot N(p), 0) \, dw_i^{'}},
\end{split}
\end{equation}
where $\Omega$ is the hemisphere outside the object surface and $\Omega^{'}$ is the hemisphere inside the object. $L(w_i)$ stands for the incident light radiance that comes from outside of the object and $L(w_i^{'})$ is the incident light radiance that comes from inside of the object. $f_r$ and $f_t$ are the reflectance term and transmission term of the microfacet BSDF \cite{walter2007microfacet} respectively. $R$ is the roughness map and $N$ is the surface normal map. The radiance $L(w_i^{'})$ is the result of multiple scattering through the volume before going out of the surface at $p$. We use the Radiative Transport Equation as our homogeneous subsurface scattering model:
\begin{equation}
\begin{split}
(w_o^{'} \cdot \nabla)L(w_o^{'}) = & -\sigma_t L(w_o^{'})\\
& + \sigma_s \int_{S^2} L(w_i^{'}) f_p(w_i^{'}, w_o^{'}, g)\, dw_i^{'} ,
\end{split}
\end{equation}
where $L(w_i^{'})$ and $L(w_o^{'})$ are the incident and outgoing light radiance respectively. The integral domain $S^2$ is a sphere. $\sigma_t$ is the extinction coefficient, $\sigma_s$ is the scattering coefficient. $f_p$ is the Henyey-Greenstein phase function, it has one parameter $g$ which defines whether the scattering is forward ($g>0$), backward ($g<0$), or isotropic ($g=0$):
\begin{equation}
    f_p(\theta, g)=\frac{1}{4\pi} \frac{1-g^2}{(1+g^2-2g \cos{\theta})^{3/2}},
\end{equation}
where $\theta$ is the angle between $w_i^{'}$ and $w_o^{'}$. In the main paper, we estimate the volumetric albedo $\alpha$ which is defined as:
\begin{equation}
\alpha = \sigma_s / \sigma_t.
\end{equation}
\section{Model Details}
\label{sec:model_detail}

We assume that the size of input images is $256\times256$. All pixels and physical parameters including extinction coefficient $\sigma_t$, volumetric albedo $\alpha$, phase function parameter $g$, flashlight intensity $i$ are normalized to $-1$ to $1$. The network parameters are initialized by Normal Initialization with the mean equal to $0$ and variance equal to $0.02$. During the training, we set batch size to $32$. We train the model for 20 epochs by using Adam optimizer \cite{kingma2014adam} with $\beta_1$ = 0.5, $\beta_2$ = 0.999, and learning rate $lr$ = 0.0002 in the first 10 epochs and a linear decay in the remaining 10 epochs. We also use Batch Normalization to stabilize training. For the loss functions, we empirically set the weight of depth $L_D$ to $5$ and the others to $1$.

\section{Network Structure}
\label{sec:network_structure}
We illustrate the detailed network structure in this section. The estimator contains a encoder and several heads. We show the structure of the proposed encoder in Table \ref{tab:encoder}, Normal, Roughness, and Depth head in Table \ref{tab:head}, Scattering and Illumination head in Table \ref{tab:sss_head}. Our neural renderer contains a Surface Encoder, a Scattering Encoder and a Decoder, we show the structure in Table \ref{tab:surface_encoder}, \ref{tab:scattering_encoder}, and \ref{tab:decoder}, respectively.

\begin{table*}[t]
\centering
\caption{Network structure of the proposed encoder. Numbers in the building blocks mean kernel size, number of output channel, and stride, respectively.}
\begin{tabular}{*{3}c}
    \toprule
     stage & building blocks & output size \\
    \midrule
    input convolution & $ \begin{bmatrix}
7\times 7 & 64 & 1 \times 1 \\
 & \text{BN} &  \\
 & \text{ReLU} &  \\
\end{bmatrix}  $ & $H \times W \times 64$\\
    \midrule
    downsampling convolution 1 & $ \begin{bmatrix}
3\times 3 & 128 & 2 \times 2 \\
 & \text{BN} &  \\
 & \text{ReLU} &  \\
\end{bmatrix}  $ & $\frac{H}{2} \times \frac{W}{2} \times 128$\\
    \midrule
    downsampling convolution 2 & $ \begin{bmatrix}
3\times 3 & 256 & 2 \times 2 \\
 & \text{BN} &  \\
 & \text{ReLU} &  \\
\end{bmatrix}  $ & $\frac{H}{4} \times \frac{W}{4} \times 256$\\
    \midrule
    resnet blocks & $ \begin{bmatrix}
3\times 3 & 256 & 1 \times 1 \\
 & \text{BN} &  \\
 & \text{ReLU} &  \\
 3\times 3 & 256 & 1 \times 1 \\
 & \text{BN} &  \\
 & \text{ReLU} &  
\end{bmatrix}  \times 9$  & $\frac{H}{4} \times \frac{W}{4} \times 256$\\

    \bottomrule
  \end{tabular}
\label{tab:encoder}
\end{table*}

\begin{table*}[t]
\centering
\caption{Network structure for the Normal, Roughness, and Depth head. Numbers in the building blocks mean kernel size, number of output channel, and stride, respectively. Output channel $C$ for Normal head is 3, and for Roughness and Depth head is 1.}
\begin{tabular}{*{3}c}
    \toprule
     stage & building blocks & output size \\
    
    \midrule
    upsampling convolution 1 & $ \begin{bmatrix}
3\times 3 & 128 & 2 \times 2 \\
 & \text{BN} &  \\
 & \text{ReLU} &  \\
\end{bmatrix}  $ & $\frac{H}{2} \times \frac{W}{2} \times 128$\\
    \midrule
    upsampling convolution 2 & $ \begin{bmatrix}
3\times 3 & 64 & 2 \times 2 \\
 & \text{BN} &  \\
 & \text{ReLU} &  \\
\end{bmatrix}  $ & $H \times W \times 64$\\
    \midrule
    output convolution & $ \begin{bmatrix}
7\times 7 & C & 1 \times 1 \\
 & \text{BN} &  \\
\end{bmatrix}  $ & $H \times W \times C$\\
    
    \bottomrule
  \end{tabular}
\label{tab:head}
\end{table*}

\begin{table*}[t]
\centering
\caption{Network structure for the Scattering and Illumination head. Numbers in the building blocks mean kernel size, number of output channel, and stride, respectively. Output channel $C$ for SSS head is 7, and for Illumination head is 28.}
\begin{tabular}{*{3}c}
    \toprule
     stage & building blocks & output size \\
    \midrule
    upsampling convolution 1 & $ \begin{bmatrix}
3\times 3 & 128 & 2 \times 2 \\
 & \text{BN} &  \\
 & \text{ReLU} &  \\
\end{bmatrix}  $ & $\frac{H}{2} \times \frac{W}{2} \times 128$\\
    \midrule
    upsampling convolution 2 & $ \begin{bmatrix}
3\times 3 & 64 & 2 \times 2 \\
 & \text{BN} &  \\
 & \text{ReLU} &  \\
\end{bmatrix}  $ & $H \times W \times 64$\\
    \midrule
    output linear & $ \begin{bmatrix}
 & C &  \\
 & \text{BN} &  \\
\end{bmatrix}  $ & $ C$\\
    
    \bottomrule
  \end{tabular}
\label{tab:sss_head}
\end{table*}

\begin{table*}[t]
\centering
\caption{Network structure for the Scattering Encoder. Numbers in the building blocks mean kernel size, number of output channel, and stride, respectively.}
\begin{tabular}{*{3}c}
    \toprule
     stage & building blocks & output size \\
    \midrule
    upsampling convolution 1 & $ \begin{bmatrix}
4\times 4 & 1024 & 2 \times 2 \\
 & \text{BN} &  \\
 & \text{ReLU} &  \\
\end{bmatrix}  $ & $\frac{H}{128} \times \frac{W}{128} \times 1024 $\\
    \midrule
    upsampling convolution 2 & $ \begin{bmatrix}
4\times 4 & 512 & 2 \times 2 \\
 & \text{BN} &  \\
 & \text{ReLU} &  \\
\end{bmatrix}  $ & $\frac{H}{64} \times \frac{W}{64} \times 512 $\\

\midrule
    upsampling convolution 3 & $ \begin{bmatrix}
4\times 4 & 256 & 2 \times 2 \\
 & \text{BN} &  \\
 & \text{ReLU} &  \\
\end{bmatrix}  $ & $\frac{H}{32} \times \frac{W}{32} \times 256 $\\

\midrule
    upsampling convolution 4 & $ \begin{bmatrix}
4\times 4 & 128 & 2 \times 2 \\
 & \text{BN} &  \\
 & \text{ReLU} &  \\
\end{bmatrix}  $ & $\frac{H}{16} \times \frac{W}{16} \times 128 $\\

\midrule
    upsampling convolution 5 & $ \begin{bmatrix}
4\times 4 & 64 & 2 \times 2 \\
 & \text{BN} &  \\
 & \text{ReLU} &  \\
\end{bmatrix}  $ & $\frac{H}{8} \times \frac{W}{8} \times 64 $\\

\midrule
    upsampling convolution 6 & $ \begin{bmatrix}
4\times 4 & 32 & 2 \times 2 \\
 & \text{BN} &  \\
 & \text{ReLU} &  \\
\end{bmatrix}  $ & $\frac{H}{4} \times \frac{W}{4} \times 32 $\\
    
    \bottomrule
  \end{tabular}
\label{tab:scattering_encoder}
\end{table*}

\begin{table*}[t]
\centering
\caption{Network structure of the Surface Encoder. Numbers in the building blocks mean kernel size, number of output channel, and stride, respectively.}
\begin{tabular}{*{3}c}
    \toprule
     stage & building blocks & output size \\
    \midrule
    input convolution & $ \begin{bmatrix}
7\times 7 & 64 & 1 \times 1 \\
 & \text{BN} &  \\
 & \text{ReLU} &  \\
\end{bmatrix}  $ & $H \times W \times 64$\\
    \midrule
    downsampling convolution 1 & $ \begin{bmatrix}
3\times 3 & 128 & 2 \times 2 \\
 & \text{BN} &  \\
 & \text{ReLU} &  \\
\end{bmatrix}  $ & $\frac{H}{2} \times \frac{W}{2} \times 128$\\
    \midrule
    downsampling convolution 2 & $ \begin{bmatrix}
3\times 3 & 256 & 2 \times 2 \\
 & \text{BN} &  \\
 & \text{ReLU} &  \\
\end{bmatrix}  $ & $\frac{H}{4} \times \frac{W}{4} \times 256$\\
    
    \bottomrule
  \end{tabular}
\label{tab:surface_encoder}
\end{table*}

\begin{table*}[t]
\centering
\caption{Network structure of the Decoder. Numbers in the building blocks mean kernel size, number of output channel, and stride, respectively.}
\begin{tabular}{*{3}c}
    \toprule
     stage & building blocks & output size \\

    \midrule
    resnet blocks & $ \begin{bmatrix}
3\times 3 & 288 & 1 \times 1 \\
 & \text{BN} &  \\
 & \text{ReLU} &  \\
 3\times 3 & 288 & 1 \times 1 \\
 & \text{BN} &  \\
 & \text{ReLU} &  
\end{bmatrix}  \times 9$  & $\frac{H}{4} \times \frac{W}{4} \times 256$\\
    
    \midrule
    upsampling convolution 1 & $ \begin{bmatrix}
3\times 3 & 128 & 2 \times 2 \\
 & \text{BN} &  \\
 & \text{ReLU} &  \\
\end{bmatrix}  $ & $\frac{H}{2} \times \frac{W}{2} \times 128$\\
    \midrule
    upsampling convolution 2 & $ \begin{bmatrix}
3\times 3 & 64 & 2 \times 2 \\
 & \text{BN} &  \\
 & \text{ReLU} &  \\
\end{bmatrix}  $ & $H \times W \times 64$\\
    \midrule
    output convolution & $ \begin{bmatrix}
7\times 7 & 3 & 1 \times 1 \\
 & \text{BN} &  \\
\end{bmatrix}  $ & $H \times W \times 3$\\
    \bottomrule
  \end{tabular}
\label{tab:decoder}
\end{table*}
{\small
\bibliographystyle{ieee_fullname}
\bibliography{egbib}
}